\documentclass[english]{pfia}

\usepackage{amsmath}
\usepackage{amssymb}
\usepackage{authblk}
\usepackage{bm}
\usepackage{dsfont}
\usepackage{graphicx}
\usepackage{xcolor}
\usepackage{enumitem}
\usepackage{multicol}

\title{\textbf{A Fair Classifier Embracing Triplet Collapse}}

\author[1]{Alice Martzloff}
\author[2]{Nicolas Posocco}
\author[1]{Quentin Ferré}
\affil[1]{Euranova, Marseille, France}
\affil[2]{Euranova, Mont-Saint-Guibert, Belgique}
\date{firstname.lastname@euranova.eu}

\begin{document}

\maketitle


\begin{resume}
Dans ce papier, nous étudions les comportements de la fonction de coût par triplet et montrons que ceux-ci peuvent être exploités pour limiter les biais crées et perpétués par les modèles d'apprentissage automatique. 
Notre classifieur équitable utilise le collapse de la fonction de coût par triplet, qui nécessite une marge supérieure à la distance maximum entre deux points dans l'espace latent et une sélection stochastiques des triplets.
\end{resume}

\begin{motscles}
Équité, apprentissage métrique, fonction de coût par triplet, collapse, classifieur équitable, IA responsable.
\end{motscles}

\begin{abstract}
In this paper, we study the behaviour of the triplet loss and show that it can be exploited to limit the biases created and perpetuated by machine learning models. 
Our fair classifier uses the collapse of the triplet loss when its margin is greater than the maximum distance between two points in the latent space, in the case of stochastic triplet selection.
\end{abstract}

\begin{keywords}
Fairness, deep metric learning, triplet loss, collapse, fair classifier, responsible AI.
\end{keywords}


\section{Introduction}

Fairness is an important component of responsible AI frameworks and is a growing concern, especially from social and regulatory perspectives. It broadly aims at avoiding the learning of historical biases, such as those based on race or gender.
We consider the problem of building a fair representation of data, so that sensitive features (i.e. race, gender) cannot be recovered from the representation. To achieve this, deep metric learning is a potentially interesting approach since it builds an embedding space with desirable characteristics.
Since the behaviour of associated triplet learning is not completely understood as of today, we propose new insights on the behaviour of the loss and use this new knowledge to create an embedder into a fair embedding space.\\

Our main contributions are:
\begin{itemize}
    \item Theoretical and experimental evidence to produce knowledge on triplet learning
    \item A novel solution for learning representations that are fair and with high utility using triplet loss collapse
\end{itemize}

\section{Related Works}

Fairness is all about the non-discrimination of populations. This means avoiding the creation or reinforcement of human prejudice and promoting inclusion and diversity. This task is difficult in practice, since datasets with supervision carry the historical annotation bias. 


The scientific community proposed metrics for the study of various definitions of fairness like Demographic parity \cite{dwork2012fairness}, Equality of opportunity \cite{hardt2016equality} and Counterfactual fairness \cite{kusner2017counterfactual}.
After observing that some models were unfair, special efforts have been allocated to finding ways to mitigate those biases in the ML pipeline, with pre-processing approaches \cite{barocas2017fairness, zemel2013learning}, which impact the data before it is ingested by the model, post-processing approaches \cite{kamiran2010discrimination}, which straighten out the model's output, and in-processing approaches \cite{zafar2017fairness, friedler2019comparative}, which get into the core of the model to make a difference.
Collecting more representative data is most of the time either impossible or too costly. As a result, the issue is usually tackled by processing existing data. To that end, pre-processing methods aim at creating a new representation of the data that is fair with respect to one of the many fairness definitions. 
It was first tried using rough baselines \cite{kamiran2012data, feldman2015certifying}, and, more recently, variational autoencoders \cite{louizos2015variational, locatello2019fairness, creager2019flexibly} and modern generative models \cite{balunovic2021fair}. In this very last publication, Balunovic et al. explain how most fair representation still allow discrimination based on the sensitive feature, in the learned representation.

Metric learning for fairness \cite{ilvento2019metric, kulis2013metric} aims at achieving individual fairness, which means that the embeddings of similar samples (except for the sensitive feature) should be very close.
Recent papers like the one of Shen et al. \cite{shen2021contrastive} already looked into the potential of contrastive learning for fairness. 
Quite unexplored yet in this context, we focus on triplet learning. 
First used by \cite{schroff2015facenet} for computer vision, Hermans et al. \cite{hermans2017defense} re-argue its arbitrary capacity as a key component for outperforming other methods.

The use of metric learning has increased since its rise in computer vision, but few had the objective of in-depth studying the behaviours related to the use of triplet learning. 
Some \cite{wang2017normface} try to explain the "tricks" to favour a quality training like adding a softmax activation function on the last layer of the embedder. Others have shed some light on collapse modes. 
A \emph{generalised collapse} (all embeddings are projected onto a single point) can occur when using the triplet loss \cite{reddit} with a hard or semi-hard selection of triplets \cite{schroff2015facenet}. 
An \emph{intra-class collapse} would group all the data points of the same class into one point \cite{levi2020reducing} if the chosen margin allows the optimisation to reach that state \cite{hermans2017defense}.

We use supervised triplet learning with several different triplet selection methods to create representations where similar samples have similar representations \cite{ilvento2019metric, kulis2013metric}, and from which the target feature can be recovered (downstream task) but not the sensitive feature(s).

To do so, we study and propose an explanation of the following phenomena: Why does the activation function enable training? 
How does it interact with the triplet loss margin? What is the link between triplet selection and collapsing? And how to use it to enforce fairness constraints?




\section{Problem Statement}

Let's focus on a binary supervised classification task. $\mathcal{D} =\{(x, y)\}$ is a set of samples from $\bm{X} \times \bm{Y}$ where $x = [x_1, ..., x_K, s]$ is a features vector and $y$ a binary label. $s \in \{0, 1\}$ is a binary sensitive feature.
In this paper, we propose to learn an embedder $\mathcal{E}$ into a $d$-dimensional fair embedding space $\bm{Z}$. $z = \mathcal{E}(x)$ denotes the embedding representation.

We chose a criterion for measuring fairness that relies on the accuracy of recovering the sensitive feature. We measure how much any adversary can recover sensitive information from the embedding. By doing so, we want to ensure that no sensitive information is available and therefore cannot contribute to the decision. This should cover any type of harm resulting from the use of a dataset.

We consider that an embedding is fair if for any embedding $z \in \bm{Z}$, it is impossible to retrieve the corresponding sensitive feature $s$, while $z$ remains a good predictor for $y$. The fairness of the embedding is verified through a supervised classification algorithm that:
\begin{itemize}[nolistsep,noitemsep]
    \item shall achieve a much better performance (e.g. area under the ROC) on the task of predicting $s$ from $x_{1}, ..., x_{K}$ than from $z$ ;
    \item shall achieve a similar performance on the task of predicting $y$ from $x$ or from $z$.
\end{itemize}

In the following, we use a deep metric learning approach to compute such a fair embedder. 

\section{Proposed Solution}

We propose to train a small multi-layer perceptron using a triplet loss, so that the \emph{selection of triplets} induces desirable properties in the latent space with regard to fairness.\\

Given a triplet $((x^a,y^a), (x^+,y^+), (x^-,y^-)) \in \mathcal{D}^3$ (anchor, positive, negative), $(z^a, z^+, z^-)$ are the corresponding embeddings. With $\alpha \in \mathbb{R}$ the triplet margin, and assuming the use of the Euclidian distance, the optimised loss is:
$$
\mathcal{L} = max\left( \| z^a - z^+ \|_2^2 - \| z^a - z^- \|_2^2 + \alpha , 0\right)
$$

\paragraph{Benchmarked triplet selection methods.}

We investigate five different triplet selection methods. The first three are studied as potential bias mitigation techniques. The latter two are control baselines meant to support our hypotheses made on the expected behaviours of the first three methods. 
\begin{enumerate}

\item \emph{classical}: $x^+$ is selected at random so that $y^+ = y^a$ (the anchor can't be re-selected). $x^-$ is selected randomly amongst samples for which $y^- \neq y^a$.

\item \emph{counterfactual}: the positive is a copy of the anchor but with the sensitive feature flipped, thus $x^+ = [x_1^a, ..., x_K^a, \overline{s}]$. $x^-$ is selected randomly amongst samples for which $y^- \neq y^a$.

\item \emph{target-agnostic counterfactual}: $x^+ = [x_1^a, ..., x_K^a, \overline{s}]$.  $x^-$ is selected randomly among all samples except the anchor.

\item \emph{random}: Both $x^+$ and $x^-$ are selected randomly among all samples but the anchor.

\item \emph{identical positive}: The positive is a copy of the anchor, thus $x^+ = x^a$. $x^-$ is selected at random amongst samples for which $y^- \neq y^a$. 

\end{enumerate}

Note that we don't use hard or semi-hard selection of triplets which has $O(n)$ time complexity, but a stochastic selection of triplets which has time complexity of $O(1)$.

\paragraph{Benchmarked latent activations/normalisations.}

We also study the influence of the use of different activations/normalisations of the output layer of the MLP embedder, used to produce the embedding. Inspired by the existence of a "softmax trick" mentioned in many papers \cite{wang2017normface}, we look into the following activations/normalisations: no activation, batch normalisation, L1 normalisation, L2 normalisation, sigmoïd, softmax and hyperbolic tangent, and study their performance qualitatively and quantitatively.

\begin{figure*}
\begin{center}
\includegraphics[width=\textwidth]{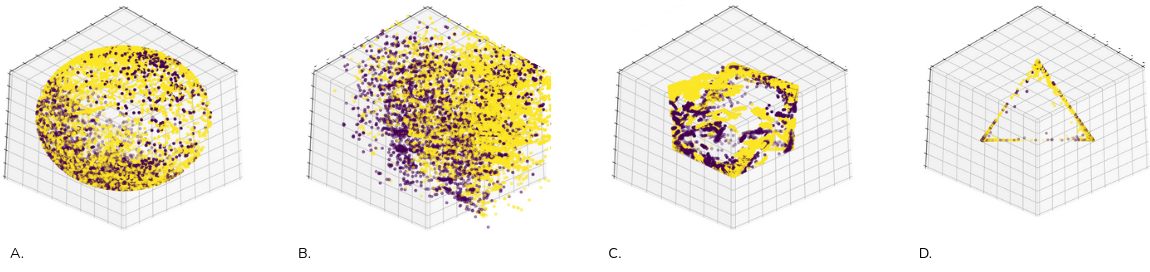}
\end{center}
\caption{State of the embedding manifold at initialization, for L2 normalisation (A), with no activation (B), for L1 normalisation (C), for softmax activation (D). Bi-coloration helps distinguish between different values for the sensitive feature.}
\end{figure*}

\section{Understanding Triplet Learning}

In this section, we discuss the expected behaviour of the triplet loss based on a theoretical analysis.\\


\paragraph{The activation shapes the embedding space.}

We apply an activation function on the embedding defined by our embedder. This operator determines the shape of the embedding space, as well as the maximum distance between two samples.\\

For the softmax activation, for instance, the embedding space is a $d$-dimensional simplex, thus the maximum distance between two points is $\sqrt{2}$. A sigmoid activation restricts the embedding space to a $d$-dimensional hypercube, thus the maximum distance is $\sqrt{3}$ if $d=3$.\\

Using the same idea and the same $d$, one can prove that for hyperbolic tangent, L1 normalisation and L2 normalisation, the maximum euclidean distances are respectively $2\sqrt{3}$, $2$ and $2$.\\

\paragraph{A margin greater than the maximum distance prevents convergence.}

If we note: 

$$
\Lambda(z^a,z^+,z^-) =
	\begin{cases}
      	1 & \text{if } \alpha \geq  \| z^a - z^- \|_2^2 - \| z^a - z^+ \|_2^2  \\
      	0 & \text{otherwise,}
	\end{cases}
$$

then the partial derivatives of the loss with respect to each embedding component is:

$$
\frac{\partial \mathcal{L}}{\partial z^a} = 
	\begin{cases}
      	\sum^{D}_{i=1} 2(z_i^- - z_i^+) & \text{if } \Lambda(z^a, z^+, z^-) = 1\\
      	0 & \text{otherwise}
	\end{cases}
$$
$$
\frac{\partial \mathcal{L}}{\partial z^+} = 
	\begin{cases}
      	\sum^D_{i=1} -2(z_i^a - z_i^+) & \text{if } \Lambda(z^a, z^+, z^-) = 1\\
      	0 & \text{otherwise}
	\end{cases}
$$
$$
\frac{\partial \mathcal{L}}{\partial z^-} = 
\begin{cases}
  	\sum^D_{i=1} 2(z_i^a - z_i^-) & \text{if } \Lambda(z^a, z^+, z^-) = 1\\
  	0 & \text{otherwise}
\end{cases}
$$

Conceptually, the impact of the loss optimisation on the embeddings can be decomposed into two forces, whose magnitude depends on the loss gradients:

- an attractive force applied to the pair of embeddings for the anchor and the positive sample

- a repulsive force applied to the pair of embeddings for the anchor and the negative sample. 

These forces are null if and only if for one of these two pairs: either the two embeddings are distant by more than the margin, or they are exactly equal.


If the margin is zero, the optimisation goes on until positive samples are at least as close as corresponding negative samples to the anchor:
$$
\alpha = 0 \Rightarrow \quad  \Lambda = 1 \quad \text{iff.} \quad \| z^a - z^+ \|_2^2 \geq \| z^a - z^- \|_2^2 \\
$$

If the margin is positive, the optimisation instead goes on until the negative samples are farther from the anchor by $\alpha$ compared to the positive samples:
$$
\alpha > 0 \Rightarrow \quad  \Lambda = 1 \quad \text{iff.} \quad \| z^a - z^+ \|_2^2 \geq \| z^a - z^- \|_2^2 - \alpha \\
$$

If the margin is positive and above a maximal distance (which we call this an outrageous margin in the following), there will always be non-null gradients:
$$
\alpha > max_{z, z'}(\| z - z' \|_2^2) \quad \Rightarrow \quad \Lambda = 1
$$




\paragraph{A collapse happens for triplet selection with small attractive force.}

A collapsed embedding space consists of samples distributed in a small number of clusters (which we define as the intra-cluster distance being less than $10^{-5}$). We hypothesise that a collapse happens due to an imbalance between attractive and repulsive forces. As we have shown, the derivatives of the loss are proportional to the difference between embeddings components. If the embeddings of the anchor and the positive sample are close, we expect the attractive force (i.e. the result of the gradient w.r.t. to anchor and positive) to be very small. In this case, if we assume there is some distance between anchors and negatives because of the selection method, then the repulsive force will drive the optimisation step, and we hypothesize this will result in the embeddings clustering at the corners of the embedding space, far away from the center of mass.

\begin{figure}[!h]
\includegraphics[width=\linewidth]{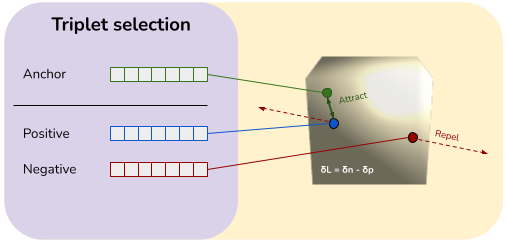}
\caption{The farther apart the embeddings in a pair are, the greater the force that is applied to them.}
\label{force}
\end{figure}

\paragraph{A random triplet selection with an outrageous margin scrambles the features.}

We hypothesise that in the case of a random triplet selection with an outrageous margin, all features will be scrambled by the collapse of the space, since no feature is targeted by the triplet loss. We define a scrambled feature as a feature that cannot be recovered from the embedding i.e. its recovery AUC is as close to 0.5 as possible.

\section{Experiments} 

\subsection{Experimental Setup}

In this section, we empirically verify previously presented theoretical insights, and show that these can be used to build fair representations in practice. We train the embedder - a simple multilayer perceptron - using a triplet approach, with previously mentioned triplet selection methods.  It projects the input feature vector $x$ into a representation with half the number of input components, and returns the embedding $z$ of output dimension 3. 

\subsection{Benchmark Datasets}

Experiments are done on the following benchmark data:
\paragraph{Adult Income Dataset.}\footnote{https://archive-beta.ics.uci.edu/ml/datasets/adult/adult.data} is a tabular dataset widely used in the modern fairness literature. It contains 48 842 samples and 15 features including the sensitive features \textit{race} and \textit{sex} to predict whether an individual makes more or less than \$50 000 a year. It contains one third of women, of whom only 10 \% are labelled as earning more than \$50 000 a year.
\paragraph{Law School Dataset.}\footnote{http://www.seaphe.org/databases/FOIA/lawschs1\_1.dta} comes from the Law School Admission Council (LSAC) Research Report \cite{wightman1998lsac}. It contains 20 798 samples and 12 features including the sensitive features \textit{race} and \textit{sex} to predict whether an individual will pass the exam to be admitted to law school. It contains 16 \% of non-white, of whom 72 \% are labelled as passing the bar exam compared to 92 \% for white individuals.\\

We consider one single sensitive feature for each dataset to produce the following results that are respectively sex and race. The ROC AUC on the task of predicting s and y from the Adult raw dataset is respectively 0.942 and 0.926. For the Law School raw, it is 0.816 and 0.907.

\subsection{Collapse \& Triplet Selection Method} 

In this section, we present the observations deduced from our current preliminary experiments. The full experimental results are not included, but we present our most salient findings in Table \ref{tab:AUCdraft}, namely the results for the best performing activations and triplet selection methods, and in the appendices.\\

As expected, we can observe clusters in the \emph{corners of the embedding space}. These clusters are not reduced to singular points and maintain some diversity.

\paragraph{When using the identical positive triplet selection}, the embedding space collapses, because of the single repulsive force induced by such a triplet selection.

\paragraph{With a target-agnostic counterfactual triplet selection}, we observe what we call a "rushed" collapse, meaning the collapse occurs from the very first epochs of the training.

\paragraph{For a random triplet selection}, the embedding spaces at different epochs are isomorphic to each other: they vary in boundaries, but maintain somewhat constant relative positions and avert collapse. This can be explained by the average resulting force exerted on each point.

\paragraph{Experiments with a counterfactual triplet selection} result in a collapse, likely due to the fact that the flip of the sensitive feature $s$ is not enough to change the embedding significantly. As such, it results in a behaviour similar to that observed with the \textbf{identical positive} triplet selection.

\paragraph{Classical triplet selection} did not collapse even after 1,000 epochs on the Adult dataset (see Figure \ref{collapse} in the Appendix). However, the histogram of all distances between embeddings shows that it is in the process of separating into two distinct clusters, and the embeddings are not spread evenly between them. This supports the idea that a \emph{classical} triplet selection induces an intra-class collapse \cite{levi2020reducing} in slow motion. The space may has been better structured because of the remaining positive force.\\

More research is needed to ensure the space has time to acquire a desired structure before the collapse. Indeed, since the collapse is not finely controlled yet, we observe some variations in the results depending on the initialization of the optimisation.

\subsection{Collapses \& Features Scrambling}

If the scrambling can be controlled so that $s$ can't be recovered from the embedding while $y$ still can, this would result in increased fairness (see next section). Here, we validate that a controlled collapse leads to the scrambling of features.

We use a the ROC-AUC of a Random Forest (RF) classifiers with 200 trees of depth 8 averaged on three different random seeds to evaluate how much a feature can be recovered from the embedding. The RF tries to predict $s$ or $y$ based on the embedding $z$. It is important to note that these models are used for evaluation only and have absolutely no impact on the training procedure.

For the Adult dataset, the highest ROC-AUC observed (i.e. the least scrambled) for both the target and the sensitive feature are of respectively 0.925 and 0.933, as we can see in Table \ref{tab:AUCdraft}. This serves as our baseline for what happens with this dataset if nothing is done to hide $s$.

\begin{table*}[t]
\centering
\caption{AUC comparison across triplet selection methods for Adult and Law School datasets. The identical positive triplet selection does not appear in this table because we do not expect a scrambling. Instead, the goal of this control experiment is to study the repartition of the samples in the embedding space.}
\label{tab:AUCdraft}
\begin{tabular}{clccrrrcl}
Dataset & Triplet Selection Method & Activation & Margin & ROC-AUC on Y & ROC-AUC on S & Epoch & Collapsed?\\
\hline
Adult & classical & batchnorm & 0 & 0.925 & 0.933 & 1000 & no \\
Adult & classical & softmax & 0 & 0.891 & 0.888 & 1000 & no \\
Adult & classical & softmax & 1 & 0.918 & 0.792 & 1000 & no \\
Adult & classical & softmax & 3 & 0.861 & 0.679 & 1000 & no \\
Adult & classical & softmax & 100 & 0.861 & 0.679 & 1000 & no \\
Adult & counterfactual & softmax & 100 & 0.792 & 0.738 & 1000 & yes \\
Adult & counterfactual & sigmoid & 100 & 0.815 & 0.716 & 1000 & yes \\
Adult & counterfactual & tanh & 100 & 0.834 & 0.735 & 1000 & yes \\
Adult & target-agnostic counterfactual & softmax & 100 & 0.677 & 0.630 & 1000 & yes \\
Adult & random & softmax & 100 & 0.870 & 0.862 & 100 & no \\
\hline
Law School & target-agnostic counterfactual & softmax & 100 & 0.533 & 0.553 & 1000 & yes \\
Law School & counterfactual & softmax & 100 & 0.6237 & 0.614 & 1000 & yes \\
Law School & classical & none & 100 & 0.953 & 0.963 & 100 & no \\
Law School & classical & sigmoid & 100 & 0.953 & 0.768 & 100 & no \\
Law School & random & softmax & 100 & 0.806 & 0.950 & 100 & no \\

\end{tabular}
\end{table*}

\paragraph{A target-preserving triplet selection induces collapse where some target information is kept whereas a target-agnostic triplet selection may induce a total scrambling.} 

In the Adult dataset, the \emph{classical} and the \emph{counterfactual} triplet selection methods have similar results in terms of AUC. They induce some drop on the target feature predictability, but most importantly a far more significant one for the sensitive feature. In the Law School dataset, the same behaviour is observed for the \emph{classical} triplet selection, but not for the \emph{counterfactual}. This is likely due to the problem being too hard on this dataset: it is here desirable to use a triplet selection that induces a collapse but tries to preserve some information on the target. This hypothesis is reinforced by the observation that on the Adult dataset that with the classical triplet selection, the sensitive feature is already scrambled at 100 epochs, which suggests that the space acquires a structure quickly for this simpler dataset. 

\paragraph{With a lower margin, the learning stops before the collapse.} We indeed observe that the AUC on the sensitive feature drops steadily while the AUC on the target feature remains similar when we progressively increase the margin until we reach an outrageous margin.

\paragraph{Collapse is needed for the total scrambling to take place.} We observe for both datasets that the sensitive and the target feature are scrambled and get poor AUC with the \emph{target-agnostic counterfactual} triplet selection. It collapses in two or more singularities.
The \emph{random} triplet selection seems to shift the shape of the embedding space as previously mentioned, yet does not scramble the features in terms of AUC.

\section{Discussion on our Solution for Learning Fair Representations}

A collapse into multiple singularities in the embedding space would result in a fair classifier, if it produces a representation from which the sensitive feature cannot be recovered, while the target feature still can. Based on the insights from the previous section, such an outcome is possible if the following conditions are met:

\begin{itemize}
\item Using an activation providing corners in the embedding space
\item Setting an outrageous margin
\item Properly selecting the triplets
\end{itemize}

One central result of this paper is that a classical triplet selection seems to scramble the sensitive feature by collapsing into multiple clusters (usually, but not always, one per class), despite the sensitive feature being completely ignored in the triplet selection process. 
This may be contrasted with the general scrambling of all features occurring in the \emph{target-agnostic counterfactual} triplet selection. Collapses tend to scramble all features. However, when using a triplet selection method that separates embeddings based on a target feature, this target feature will be less scrambled.

As such, one can build a fair classifier by engineering a collapse that protects the target while scrambling the rest of the features (like with \emph{counterfactual} or \emph{classical} triplet selections) instead of scrambling them all (like with the \emph{target-agnostic counterfactual} triplet selection).

Compared to already existing approaches, our setup has the advantage of not being adversarial: the RFs are not used in training. Thus, our AUC can be considered unbiased unlike what was highlighted in \cite{balunovic2021fair}. The many AUCs around 0.7 we observe on the sensitive feature are higher but not much higher than the one obtained by Balunovic et al. \cite{balunovic2021fair}, whose best AUC was 0.61. In any case, it is lower than the best AUC of 0.8 reached out for the adversarial methods (after correction) mentioned in the same paper. Our approach also has the advantage of being far more lightweight than Normalizing Flows-based methods.

\paragraph{A fair classifier emerges from a balance between collapse and target-preservation.}

Our best result regarding fairness is with the \emph{classical} triplet selection. We hypothesise it reduces the AUC on the sensitive feature even if it is completely ignored by the selection because the sensitive feature is scrambled. This scrambling occurs because the space is given structure too quickly: far away elements exert a greater force on each other, so that a sample won't have the opportunity to leave a cluster to join another once it has entered it. This space structure acquired early cannot be corrected.

One reason for the aforementioned phenomenon is that we randomly select valid samples for the triplets instead of using a semi-hard or hard selection. Our method ensures that all samples may be selected in a triplet and hence moved by the gradient during the entirety of the training.
Secondly, the embedding has a high ROC-AUC in the first epochs of the training for both the sensitive and the target feature regardless of the triplet selection method. Thus, it was not obvious that a target-agnostic counterfactual selection collapse would scramble those features: this information could have been preserved by splitting into clusters.
Finally, the ROC-AUC on the target is not increased in our approaches, suggesting that the learning does not simply converge to the trivial solution "embedding equals the target". It is indeed scrambling.

Essentially, the collapse induced by the outrageous margin and the repulsive force will outpace the target-preservation induced by the triplet selection.
The next research question to be answered is how to efficiently find a balance point between them.

\section{Conclusion}

We proposed triplet learning collapse as a bias-mitigation technique to solve fairness issues in machine learning, after highlighting new insights on the associated optimisation. By controlling the behaviour of the associated triplet loss function, we are able to learn fair representations, in the sense that a predictor cannot recover the sensitive attribute, while the embedding retains enough information about the target attribute to solve the downstream classification task. Such a representation is computationally cheaper than approaches based on adversarial models or normalizing flows. 

\paragraph{Future Work.}

Our highest priority is to conduct further experiments to provide convincing evidence for the insights presented in this paper. Then, our current analysis does not rely on any hyper parameter tuning, and thus performance can be further improved. Future experiments will focus on progressively increasing the optimisation strength, to wait until the space is well structured before forcing the collapse. One way to optimise would be to change the way the triplets are selected while the training is in progress. Additional efforts are also needed to validate that we can preserve any needed feature with the triplet selection. Experimenting with a control variable different than the target feature and doing a triplet selection in such a way that more than one feature is retained will strengthen our solution. That is, this work could be extended to several sensitive features if we manage to retain several control variables, despite the scrambling.

\section*{Acknowledgement}
This work was supported by the French National Research Agency in the framework of the TAUDoS project (ANR-20-CE23-0020). 

\bibliography{biblio}
\bibliographystyle{plain}
\let\oldsection\section
\clearpage
\onecolumn

\section*{Appendix}
 
\begin{figure}[!h]
\begin{center}
\includegraphics[width=\textwidth]{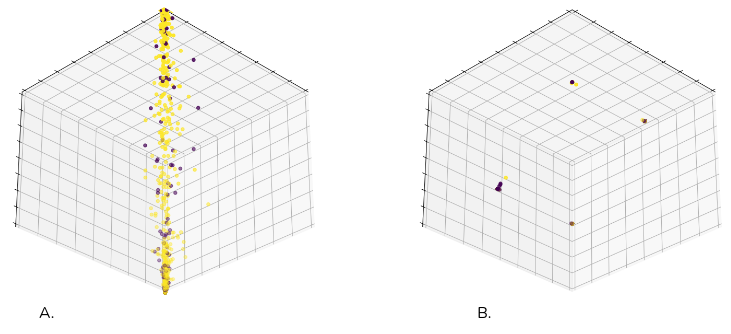}
\caption{State of the embedding manifold at epoch 500 for the Adult dataset on different seeds with an outrageous margin, for the classical triplet selection  with hyperbolic tangent activation (A), for the target-agnostic counterfactual triplet selection with sigmoid activation (B). Bi-coloration helps distinguish between different values for the sensitive feature.}
\end{center}
\label{collapse}
\end{figure}

\begin{figure}[!h]
\includegraphics[width=\textwidth]{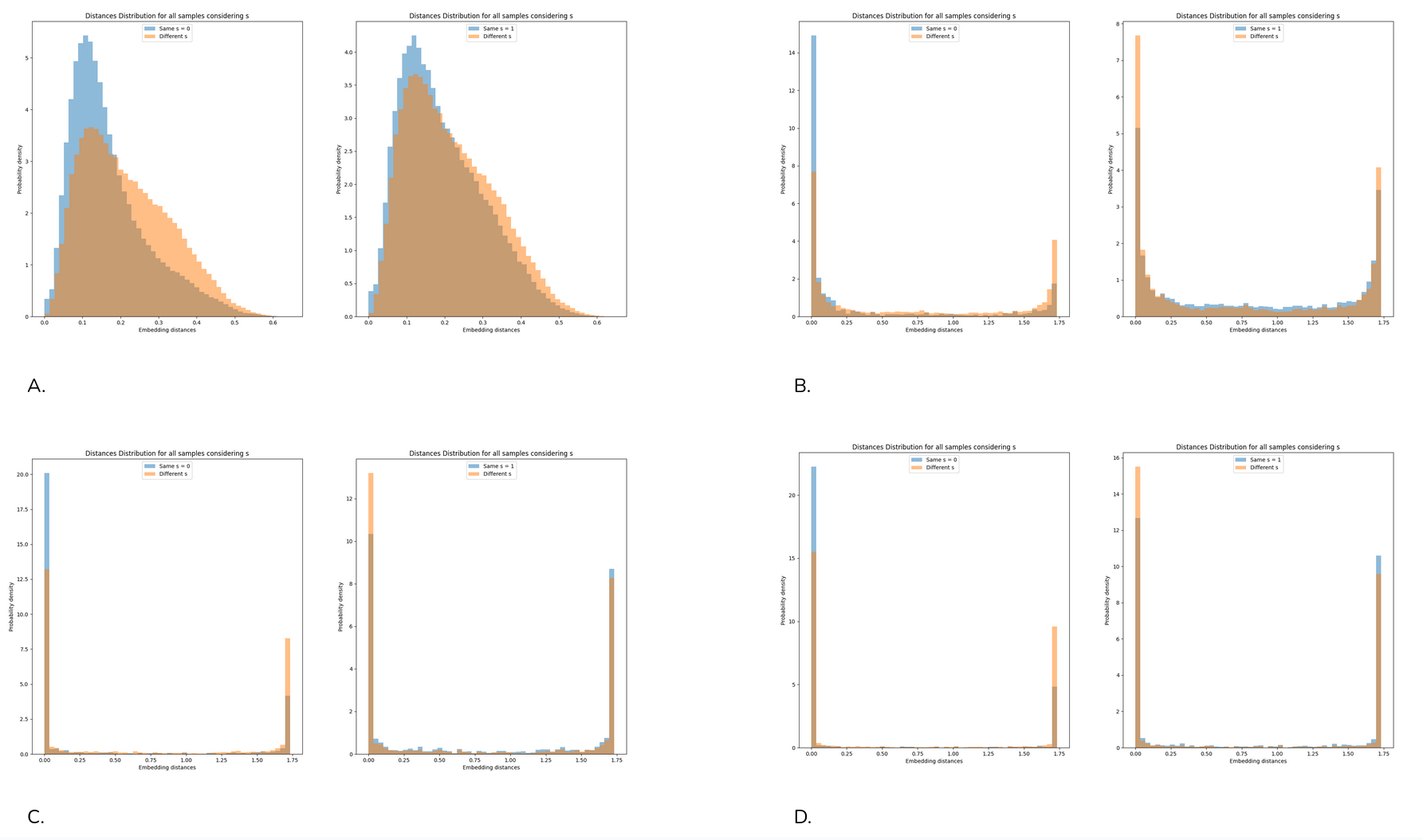}
\caption{Histogram of distances for the Adult dataset with an outrageous margin and with sigmoid activation, for the \textbf{classical} triplet selection at epoch 1 (A), epoch 10 (B), epoch 100 (C), epoch 500 (D).}
\label{baseline_distance_histogram}
\end{figure}

\begin{figure}
\includegraphics[width=\textwidth]{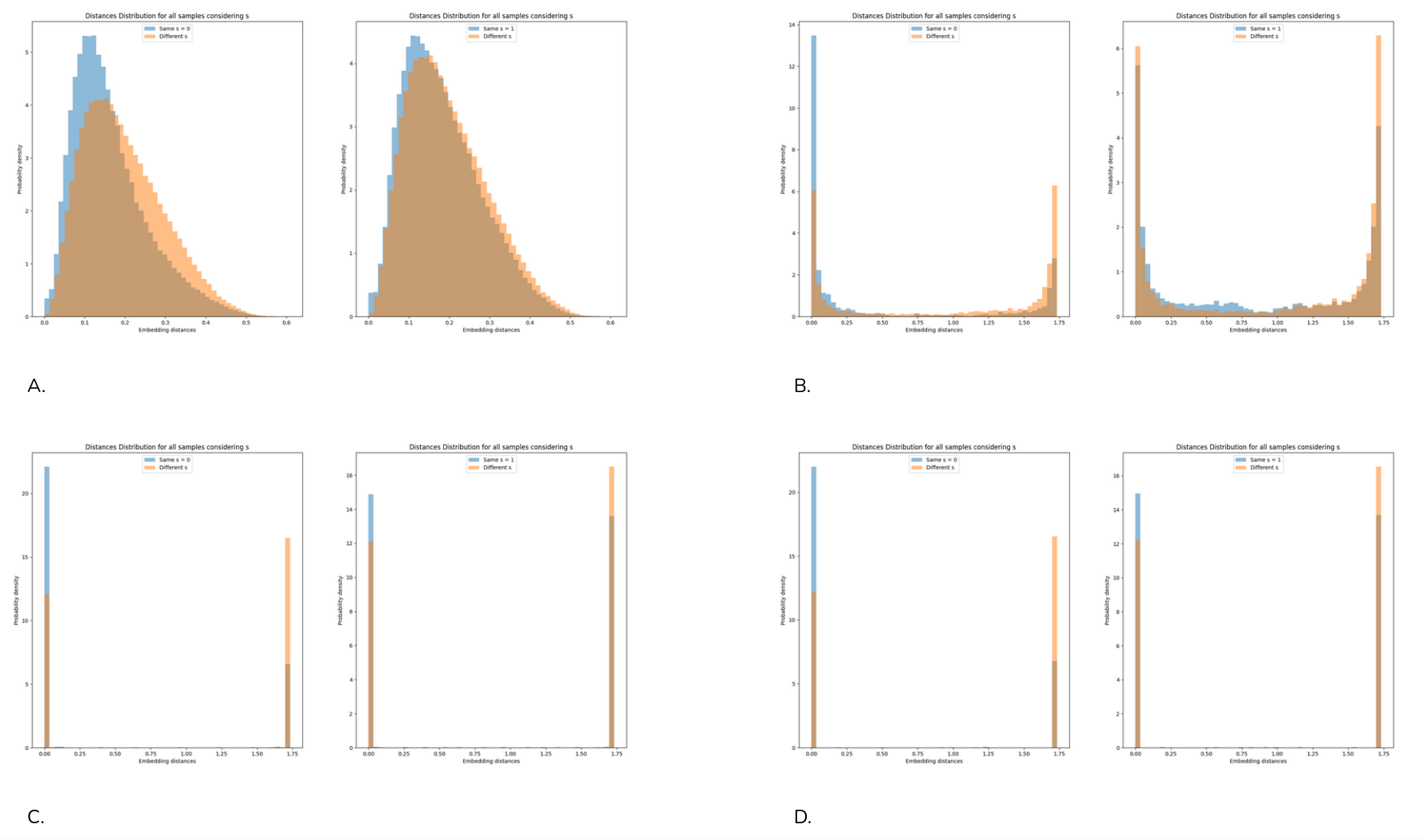}
\caption{Histogram of distances for the Adult dataset with an outrageous margin and with sigmoid activation, for the \textbf{counterfactual} triplet selection at epoch 1 (A), epoch 10 (B), epoch 100 (C), epoch 500 (D).}
\label{counterfactual_distance_histogram}
\end{figure}

\begin{figure}
\includegraphics[width=\textwidth]{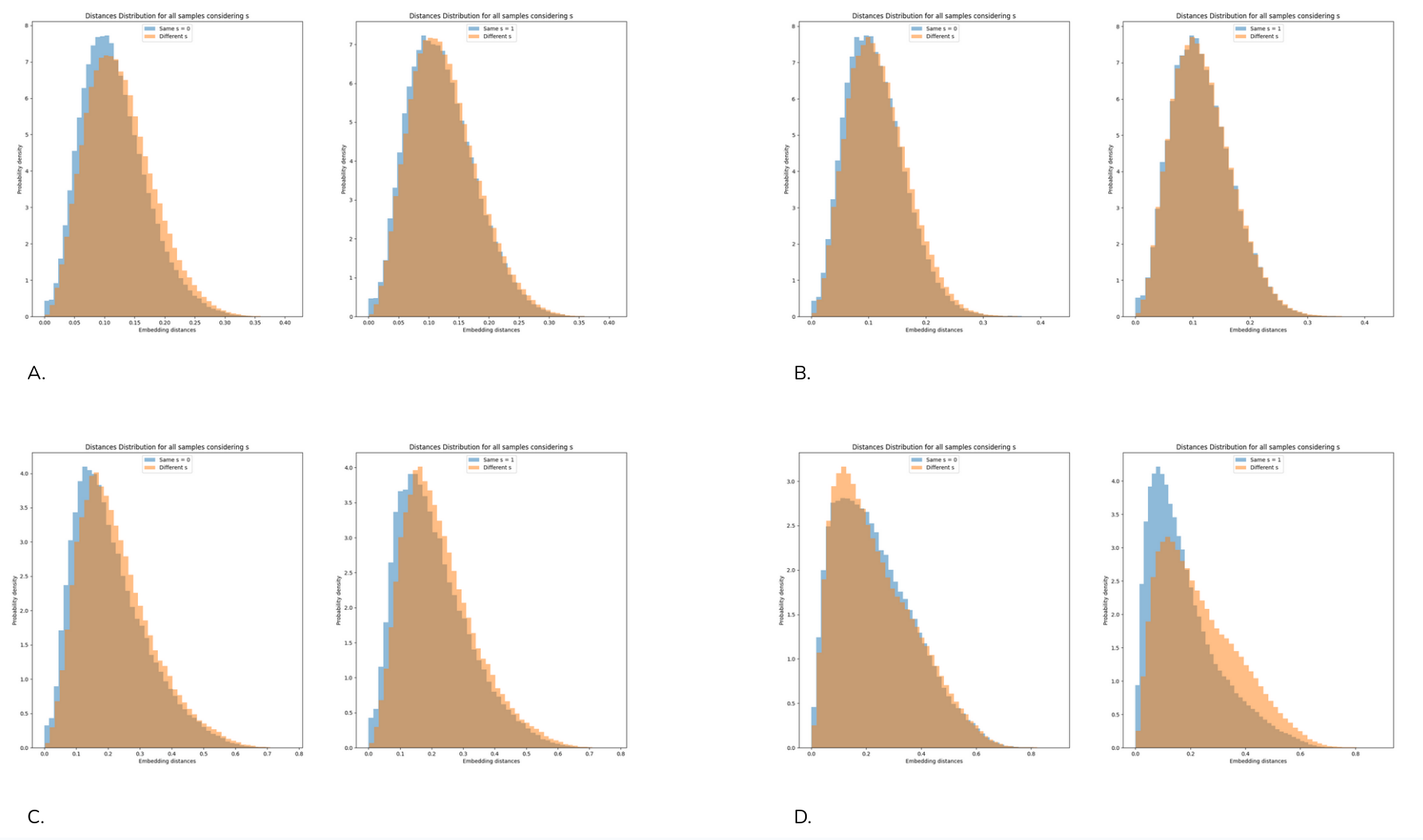}
\caption{Histogram of distances for the Adult dataset with an outrageous margin and with sigmoid activation, for the \textbf{random} triplet selection at epoch 1 (A), epoch 10 (B), epoch 100 (C), epoch 500 (D).}
\label{random_distance_histogram}
\end{figure}

\end{document}